\documentclass[sigconf]{acmart}
\renewcommand\footnotetextcopyrightpermission[1]{} 

\usepackage{booktabs} 
\usepackage{placeins}

\usepackage{algorithm}
\usepackage[noend]{algpseudocode}

\begin{document}

\title{Functional Generative Design of Mechanisms with \\ 
Recurrent Neural Networks and Novelty Search}
\renewcommand{\shorttitle}{Functional Generative Design of Mechanisms with RNNs and NS}


\author{Cameron R. Wolfe}
\affiliation{
	\institution{The University of Texas at Austin}
}
\email{wolfe.cameron@utexas.edu}

\author{Cem C. Tutum}
\affiliation{
	\institution{The University of Texas at Austin}
}
\email{tutum@cs.utexas.edu}

\author{Risto Miikkulainen}
\affiliation{
	\institution{The University of Texas at Austin}
}
\email{risto@cs.utexas.edu}

%
%

\renewcommand{\shortauthors}{C. R. Wolfe et. al.}

\begin{abstract}

Consumer-grade 3D printers have made it easier to fabricate aesthetic objects and static assemblies, opening the door to automated design of such objects. However, while static designs are easily produced with 3D printing, functional designs with moving parts are more difficult to generate: The search space is too high-dimensional, the resolution of the 3D-printed parts is not adequate, and it is difficult to predict the physical behavior of imperfect 3D-printed mechanisms. An example challenge is to produce a diverse set of reliable and effective gear mechanisms that could be used after production without extensive post-processing.  To meet this challenge, an indirect encoding based on a Recurrent Neural Network (RNN) is created and evolved using novelty search. The elite solutions of each generation are 3D printed to evaluate their functional performance on a physical test platform.  The system is able to discover sequential design rules that are difficult to discover with other methods. Compared to direct encoding evolved with Genetic Algorithms (GAs), its designs are geometrically more diverse and functionally more effective.  It therefore forms a promising foundation for the generative design of 3D-printed, functional mechanisms.

\end{abstract}



\maketitle



\section{Introduction}

\begin{figure}
\includegraphics[width=3.0in]{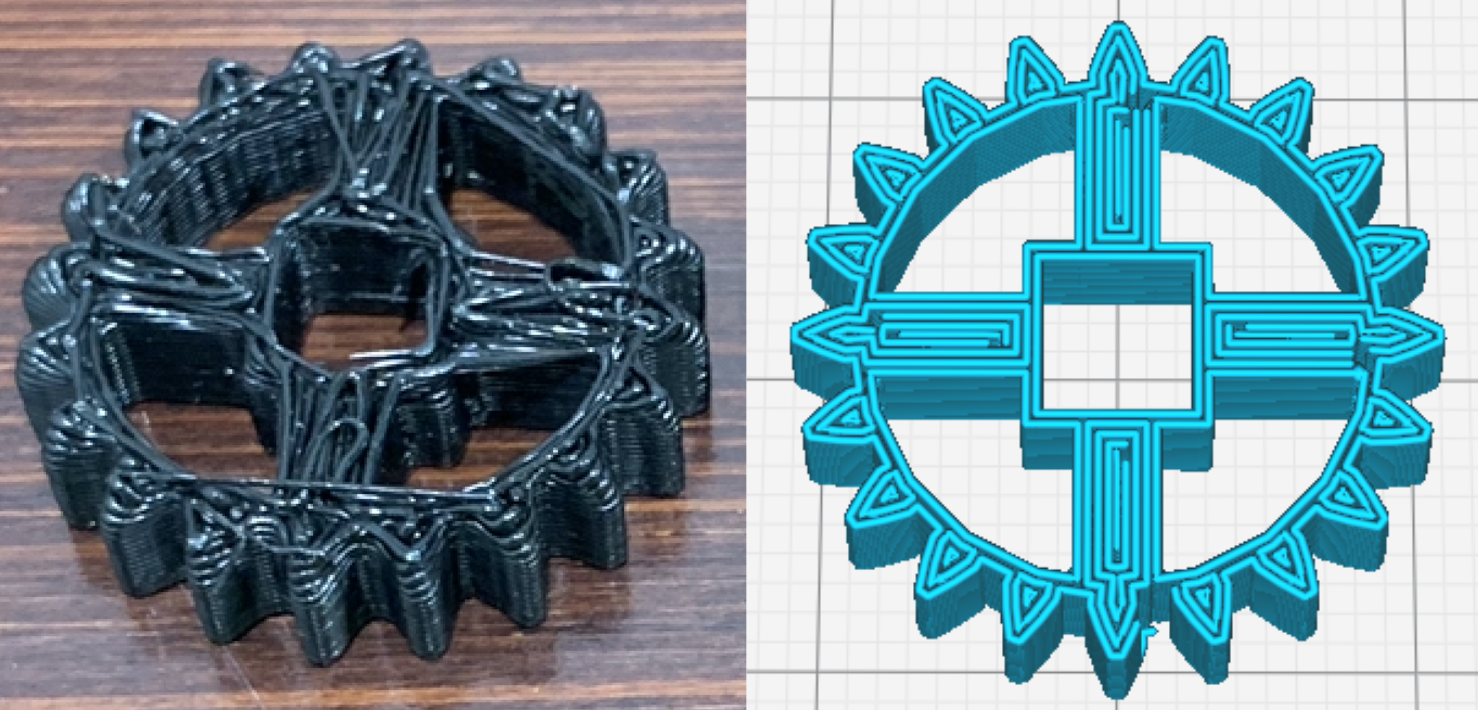}
\caption{(a) A gear that was fabricated with an FDM 3D printer and (b) the gcode file that was used to prepare the fabrication of the gear. The imprecision of FDM printers on smaller, detailed parts can be observed.}
\label{fig_gear_comparison}
\end{figure}

Fused Deposition Modeling (FDM) or commonly known as the process of 3D Printing, is a practical prototyping technique in which a 3D digital model of a design is sliced via dedicated software into thin layers and fabricated by fusing these layers on top of each other to form the final product. Its low cost and ease of use made it the standard technology for consumer-grade printers, and its wide adoption enabled fabrication of a variety of creative and aesthetic objects as well as static assemblies easier than ever. However, despite the availability of such open source designs in popular repositories for makers \cite{PLAmotor, WindupToyCar}, generation of designs with moving parts is still challenging, i.e., extensive postprocessing \cite{Gear3DPArticle, EvDecomp} or re-printing of the parts is needed. High dimensional search space for the control parameters in design (e.g., shape analysis of individual parts, their assembly for a particular functional performance, etc.) and print process (e.g., temperature and speed settings for various 3D printer parts) is one of the principal reasons. Second, slicing software inherently limits the resolution of the digital design while preparing the printing instructions as seen in Fig. \ref{fig_gear_comparison}b, where a digital gear model is discretized and discontinuities formed. Additional losses due to the nature of the printing hardware is also inevitable (see Fig. \ref{fig_gear_comparison}a) \cite{RobotGripperArticle}. Third, it is difficult to predict the physical behavior of imperfect 3D-printed mechanisms by using computer aided simulations due to numerous unknowns, i.e., material properties, stiffness and damping of parts, surface properties of the interacting parts such as friction coefficient, etc.

Design of functional mechanisms has mostly been relying on the use of pre-existing design rules, analytical equations, or user input, thus has lack of automation and emergence of creativity. There are some studies exploring the ability of cognitive models to create functional mechanisms such as wind-up toys and robots that are able to accomplish some desired tasks, such as moving in a certain direction or propelling objects. In \cite{Cor2013}, researchers proposed a semi-automated method to let the non-expert users to iteratively create animated mechanical characters. The match between the sketch of the motion curve of each part and 3D printed designs were analyzed. Another interactive design method was developed in \cite{Duy2017} which enables users to rearrange an existing mechanism to fit within a desired space and fabricate them, but it requires some experience in design. In \cite{Song2017}, a computational system was developed to construct a compact internal mechanism for wind-up toys to realize user-requested part motions utilizing a database of known mechanism elements and their corresponding motion transfer. Some recent studies focused on the design and optimization of gear mechanisms in particular. For example, multiobjective genetic algorithms were used together with a set of analytical equations to simultaneously maximize the efficiency and minimize the volume of a design having a pair of gears in \cite{Mil2018}. A similar approach was performed by \cite{Bart2018} for a larger system of gears. Gologlu and Zeyveli \cite{Gol2009} automated the preliminary stages of designing a gear system using GAs to evolve a set of possible gear mechanisms as a means of support in decision-making process, wheras Savsani et.al. \cite{Sav2010} applied particle swarm optimization algorithm to achieve a similar goal. Despite such attempts, these approaches in common utilize known analytical or emprical equations and design constraints, thus limiting less intuitive and novel mechanisms.

On the other hand, recent advances in generative models, mostly utilizing unsupervised deep-learning models but not limited to, learn the common features in vast amount of data, i.e., 2D images or 3D digital geometry files. For example, electromechanical robots, comprised of elementary building blocks (rods and joints), were autonomously designed and optimized in a study performed by Lipson and Pollack \cite{Lip2000}. This autonomous design method was successful in generating a diverse set of robot mechanisms traversing a target distance. Additionally, a Generative Adversarial Network (GAN) model was implemented to learn an efficient representation from a dataset of 3D shapes such as furnitures, chairs, etc., eventually to generate novel designs that are not existing in the training set. Although these designs were not physically fabricated, the 3D-GAN was successful in generating high-resolution 3D representations of new designs. Despite such interesting applications, the use of generative models has been limited to static designs. Recently, Tutum et.al. \cite{FGDesign}, integrated a Variational Autoencoder with a surrogate-based optimization method to generate new 3D printable springs for a car launcher mechanism with desired functionality. In this paper, an example challenge is to produce a diverse set of reliable and effective gear mechanisms that could be used after production without extensive post-processing. To meet this challenge, an indirect encoding based on a Recurrent Neural Network (RNN) is created and evolved using novelty search. The elite solutions of each generation are 3D printed and assigned a distance score based on a physical test platform.

The organization of the paper is as follows: First, the methodology is introduced followed by the details about the implementation of evolution with novelty search criterion. Next, the physical evaluation setup for mechanism testing is described. Finally, quantitative results of gear mechanisms generated by RNNs are given and compared with those produced by a baseline GA, followed by a brief discussion of the outcomes of this methodology and possible future work.

\section{Methodology} \label{sec_method}

\begin{figure}
\includegraphics[width=3.2in]{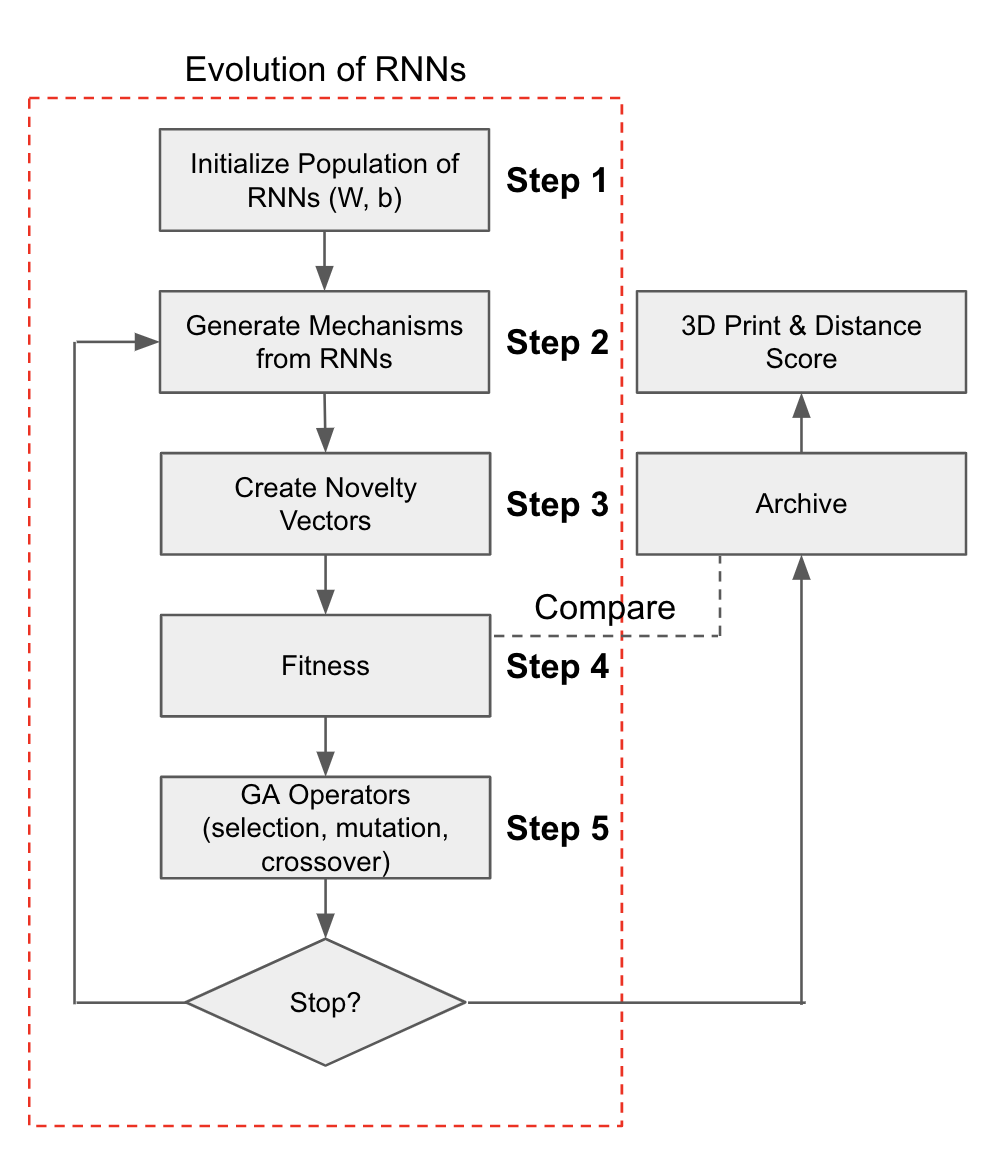}
\caption{Overall methodology for evolving RNNs that indirectly encode gear mechanisms. Evolution is driven by novelty search. The mechanism with highest novelty score is fabricated for physical evaluation during each generation.}
\label{fig_methodology}
\end{figure}

Overall methodology (see Fig.\ref{fig_methodology}) involves a Recurrent Neural Network (RNN), a novelty search algorithm, and a 3D printer to fabricate mechanism designs for functional testing (see Sec. \ref{sec_exper}). Evolution begins with a population of 150 randomly initialized RNNs, each of which encodes a single gear mechanism, and continues for 40 generations. Evolution is driven by novelty search, which assigns high fitness to RNNs that produce unique mechanisms. The generative RNN model used in this experiment designs gear mechanisms by sequentially adding gears into an empty system until a full mechanism is produced. The mechanisms with the highest novelty score in each generation are fabricated with a 3D printer and placed into an archive of elite solutions. Each mechanism that is fabricated is evaluated with a physical experiment (see Sec. \ref{sec_exper}) and assigned a distance score to better understand its functional capabilities.

Additionally, direct encodings of gear mechanisms are evolved for comparison with the evolution of generative RNN models. All aspects of the two experiments, aside from the way in which mechanisms are represented, are identical.

\newpage
\subsection{Constraint Handling for Mechanisms}

In order to generate valid gear mechanisms, some design constraints must be satisfied, such as limiting the size of the mechanism and avoiding collisions of gears within the mechanism. During evolution, the fitness of infeasible solutions (i.e., those that violate constraints) is assigned to be worse than all feasible solutions and is further penalized by the severity of the constraint violation. Therefore, the evolutionary process favors not only unique mechanisms, but also feasible mechanisms that can be fabricated. 

\begin{figure}
\includegraphics[width=3.0in]{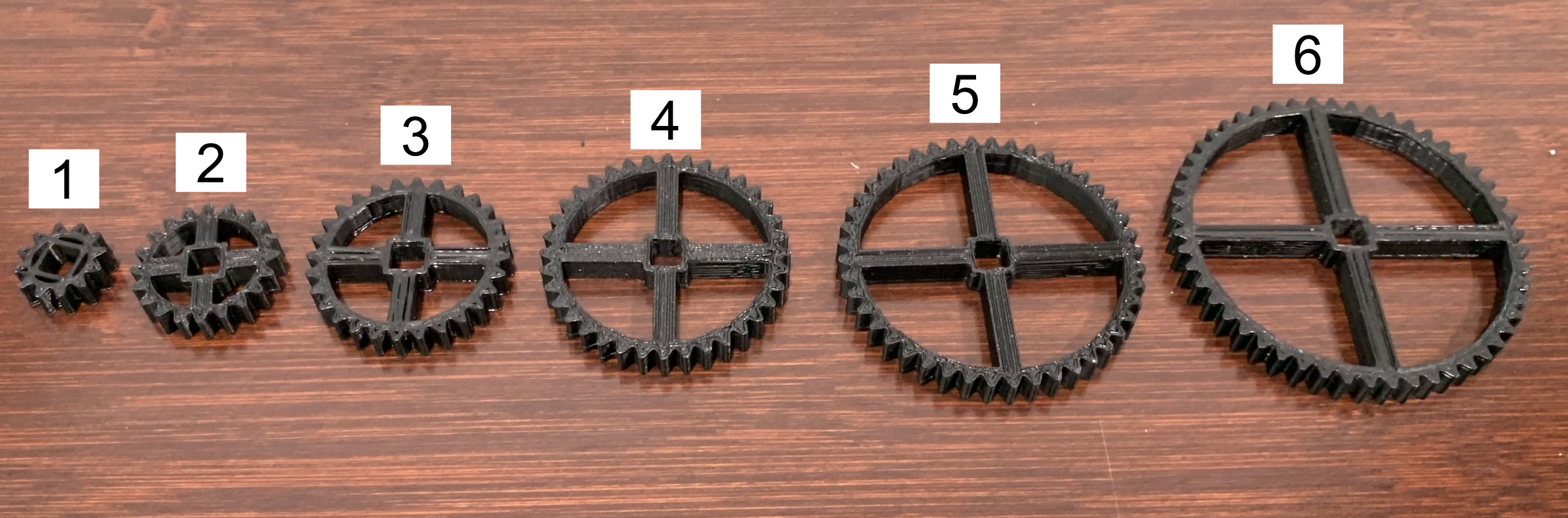}
\caption{RNN may choose to place one of these six different sizes of gears into the system at every time step.}
\label{fig_gears}
\end{figure}

\subsection{Method-I: Generative RNN Model} 

\begin{figure}
\includegraphics[width=3.5in]{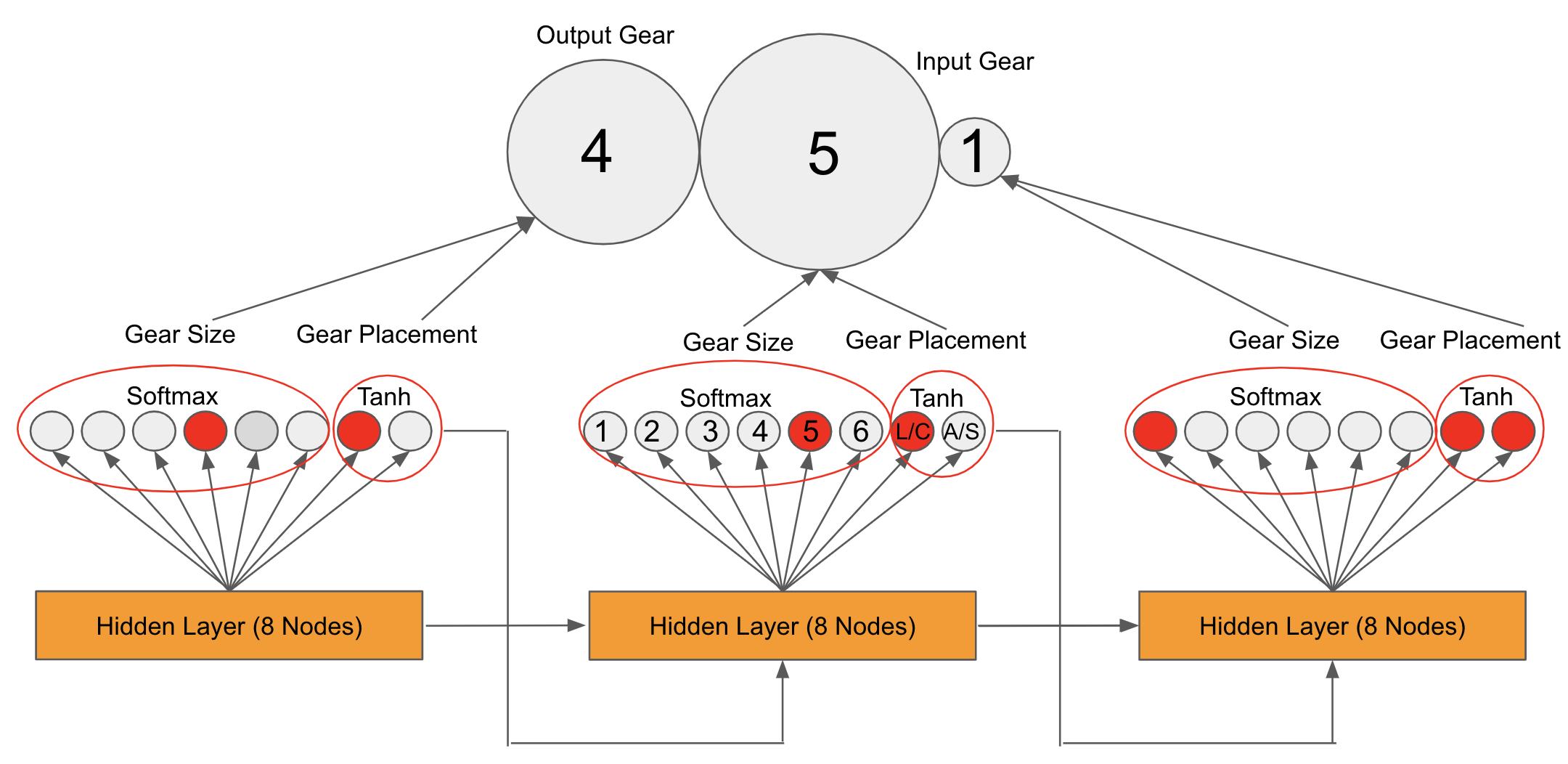}
\caption{The RNN model encodes a gear mechanism by sequentially placing gears into an empty design area. Red output nodes have high activation values, while grey nodes have low activation values. Labels 1-6 represent gear size, L/C represents linear vs. coaxial placement and A/S represents continue adding gears vs. stop adding gears.}
\label{RNN_activation}
\end{figure}

This section outlines the architecture of the generative RNN model used for the encoding of gear mechanisms, which is illustrated in Fig. \ref{RNN_activation}. The RNN used for this study has one-hidden-layer and two weight matrices (including bias), which connect the input and hidden layers to the hidden layer and hidden layer to the output layer. The input, hidden, and output layers each consist of 8 nodes.

\begin{algorithm}[H]
\caption{Evolution of RNNs}\label{RNN_flowchart}
\begin{algorithmic}[1]
\Procedure{evolve}{$rnn\_pop$}
   \State $pop\_size := 150$
   \State $n_{gen} := 40$
   \State $archive := \emptyset$
   \State $pop := random\_init\_RNN\_pop(pop\_size) \ \boldsymbol{STEP 1}$
   \State $generation := 0$
   \While{$generation<n_{gen}$}
      \State $vectors := \emptyset$
      \For {$i$ \textbf{from} $0$ \textbf{to} $pop\_size$}
          \State $output = RNN.forward(pop[i]) \ \boldsymbol{STEP 2}$
          \State $vectors.append(get\_vector(output)) \ \boldsymbol{STEP 3}$
      \EndFor
     \State $best\_individual := null$
     \State $best\_novelty := 0$
     \For{$i$ \textbf{from} $0$ \textbf{to} $pop\_size$}
         \State $fitness := distance(vectors[i], archive) \ \boldsymbol{STEP 4}$
         \State $pop[i].fitness := fitness$ 
          \If{$fitness > best\_novelty$}
             \State $best\_individual := pop[i]$
              \State $best\_novelty := fitness$
          \EndIf
      \EndFor
      \State $archive.append(best\_individual)$
      \State $pop := select\_tourn(pop)$
      \State $pop := mutation\_and\_crossover(pop) \ \boldsymbol{STEP 5}$
      \State $generation := generation+1$
    \EndWhile
\EndProcedure
    
\end{algorithmic}
\end{algorithm}

Within the network, the hidden layer contains a recursive connection, which allows the RNN to be activated over multiple time steps. During each time step, the previous output layer is used as input to the RNN. The hidden layer is obtained by concatenating the input layer with the previous hidden layer and passing this vector through a weight matrix. This hidden layer is then activated with an element-wise hyperbolic tangent activation function and passed through another weight matrix to yield the output. During the first time step, the previous hidden and output layers are initialized to vectors of ones because there is no previous hidden or output state to be used.

The RNN is executed recursively to individually place each gear into a mechanism until the mechanism is fully constructed, as can be seen in Fig. \ref{RNN_activation}. At each time step of the RNN, a single gear is added into the mechanism. This gear may be placed linear (i.e., side-by-side) or coaxial (i.e., along the same center of axis) with respect to the previous gear. A mechanism can contain a maximum of six gears and no fewer than two gears.

At each time step, the first six RNN output nodes are activated with a softmax activation function and represent the size of the gear to be placed in the mechanism. There are a total of six different gear sizes that can be added into a mechanism, as seen in Fig. \ref{fig_gears}. The same gear can be added into a mechanism multiple times. The other two output nodes are activated with a hyperbolic tangent activation function. These final two outputs determine how to place the current gear into the mechanism with respect to the previous gear (linear vs. coaxial) and whether RNN should continue adding gears into the mechanism, respectively. All output nodes, eight in total, are used as input for the following time step.  

The equations for the activation of the generative RNN model are as follows:

\begin{eqnarray}
\left.\begin{aligned}
& \boldsymbol{h_t} = \alpha(\boldsymbol{Wx_t} + \boldsymbol{Rh_{t-1}} + \boldsymbol{b_w})\\
& \boldsymbol{o_t} = \alpha(\boldsymbol{Zh_t} + \boldsymbol{b_z})   \boldsymbol{\bigcup}  \sigma(\boldsymbol{Yh_t} + \boldsymbol{b_y}) 
\end{aligned}\right.
\end{eqnarray}

where $\boldsymbol{x}$ is the input vector at time t, $\boldsymbol{h}$ is the hidden layer at time t, $\boldsymbol{b}$ is the bias vector, $\boldsymbol{W}$ is the input to hidden layer weight matrix, $\boldsymbol{R}$ is the hidden layer to hidden layer weight matrix, $\boldsymbol{Z}$ and $\boldsymbol{Y}$ are the hidden layer to output layer weight matrix (separated to illustrate two different output activation functions), $\alpha$ is the element-wise hyperbolic tangent activation function, and $\sigma$ is the softmax activation function. 

The generative RNN models are evolved for 40 generations using GA and novelty search objective. The details of evolution are outlined in section \ref{sec_evolution}.

\subsection{Method II: Direct Representation} \label{sec_GA}

In order to have a baseline model to compare with generative RNNs, a direct representation of gear mechanisms was created and evolved. This representation is comprised of an ordered list of gears for each mechanism combined with a list of placements for each gears with respect to the previous gear. A gear and placement list together encode a single mechanism, which contains between two and six total gears. Similar to RNN evolution, these direct representations were evolved for 40 generations using GA and novelty search. Aside from the way in which mechanisms are represented and created, the evolutionary process for each of these methods is identical. 

\subsection{Novelty Search} \label{sec_novelty}

\begin{figure}
\includegraphics[width=3.4in]{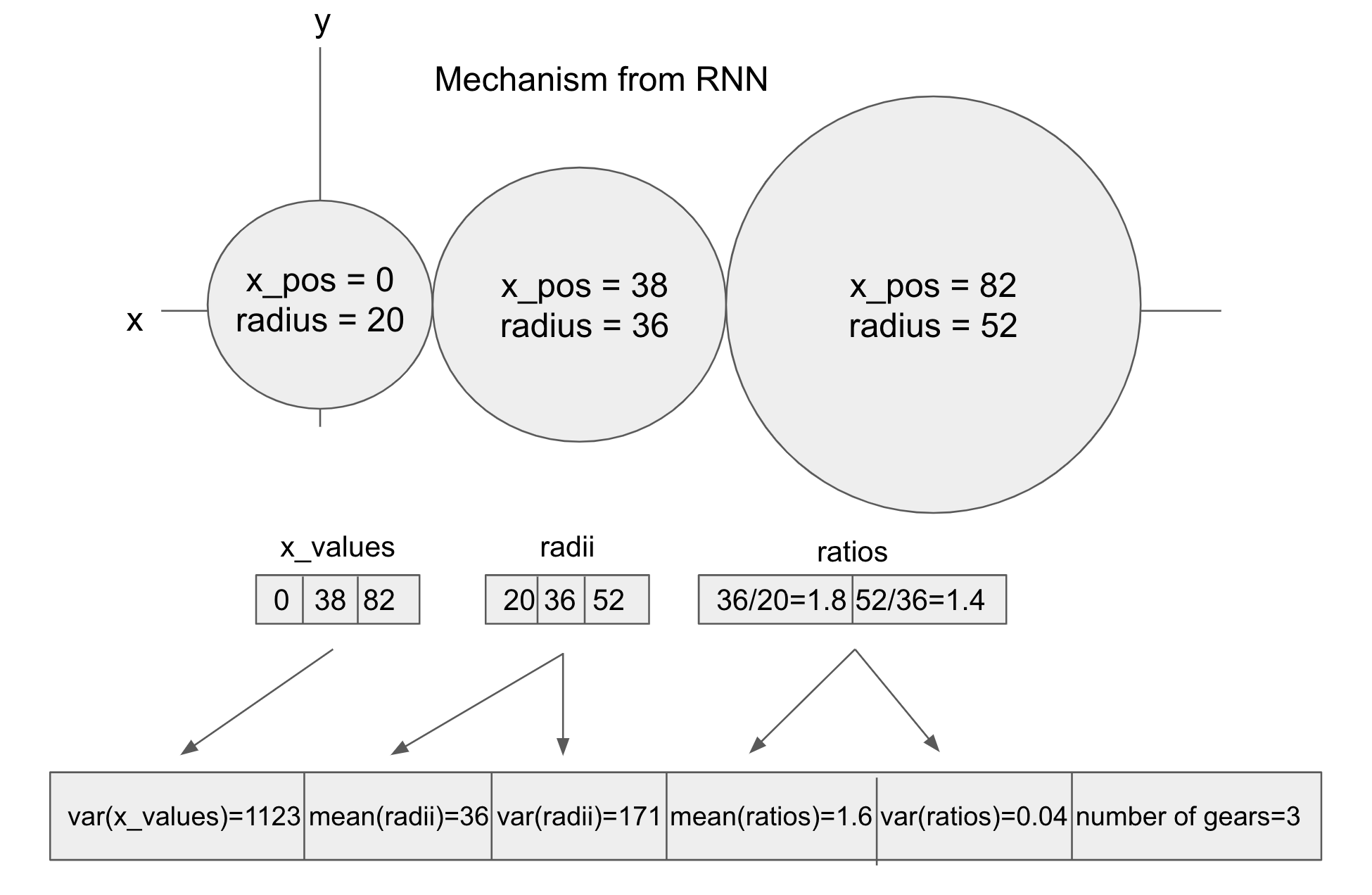}
\caption{Demonstrates how a novelty vector is created from a gear mechanism.}
\label{fig_vec}
\end{figure}

Novelty search is chosen as the objective function: \textit{i}) to allow evolution to perform well in deceptive domains, \textit{ii}) to handle the limited number of fitness evaluations due to the expense of fabricating each mechanism design. The number of feasible solutions that satisfy various design constraints is low. Therefore, searching for an effective gear mechanism would be more difficult if a specific objective was chosen (i.e., maximizing distance the car is pulled along the track). Novelty search discovers innovative designs by aiming to maximize diversity of mechanisms in the population and minimize the violation of constraints, allowing a more productive region of the large search space to be explored.

In order to assess novelty, a distance metric was created between mechanisms. For every mechanism, a vector, called the novelty vector, can be created that describes its structural properties. The values within this novelty vector include various characteristics of the mechanism, such as the ratios between gears (i.e., the quotient of radii between two adjacent gears). The vector for each mechanism includes the following values:  

\begin{itemize}
\item variance(X): Variance of x-positions of the gears (gears are placed from the front to the back of the gear box),
\item mean(ratios): Mean of gear ratios (quotient of adjacent gear radii),
\item variance(ratios): Variance in ratios (quotient of adjacent gear radii),
\item mean(radii): Mean in radii of gears put into the mechanism,
\item variance(radii): Variance in radii of gears put into the mechanism,
\item Total Number of Gears
\end{itemize}

The process of creating a novelty vector is illustrated in Fig. \ref{fig_vec}. Using this vector, novelty can be calculated by finding the Euclidean distance between novelty vectors for different mechanisms. This way of assigning fitness allowed for fitness to be determined without fabricating and testing every possible solution. Instead, only the individual with maximum novelty score after each generation fabricated for more detailed physical evaluation, thus allowing innovative, high-performing designs to be discovered without significant testing overhead.

\subsection{Evolution Process} \label{sec_evolution}

The evolution process for generative RNNs is illustrated in Fig. \ref{fig_methodology} and outlined in more detail in Algorithm \ref{RNN_flowchart}. Evolution begins by randomly initializing a population of 150 generative RNNs, each of which encodes a single mechanism. During every generation, the mechanism encoded by each RNN is generated. For every mechanism, the novelty vector is found and used to determine the mechanism's novelty score, which is assigned as the mechanism's fitness. After fitness is assigned, selection, mutation and crossover are applied to the population to create the next generation of offspring. The evolution process continues for 40 generations.

The novelty score for each mechanism, which is used to assign fitness, is calculated by finding the minimum euclidean distance between a candidate mechanism's novelty vector and all novelty vectors within an archive of elite solutions from previous generations. This archive includes the mechanisms with the highest novelty score from every generation. Moreover, each mechanism in the archive is fabricated and evaluated using the experimental setup (see Sec. \ref{sec_exper}) to determine its distance score. It should be noted, however, that this distance score is unrelated to evolution and is rather used to evaluate the functional effectiveness of mechanisms produced during evolution.

The process for evolving direct mechanism representations was identical to the evolution of RNNs, aside from the way in which mechanisms were represented and created.

\subsection{Evaluation Setup} \label{sec_exper}

\begin{figure}
\includegraphics[width=3.2in]{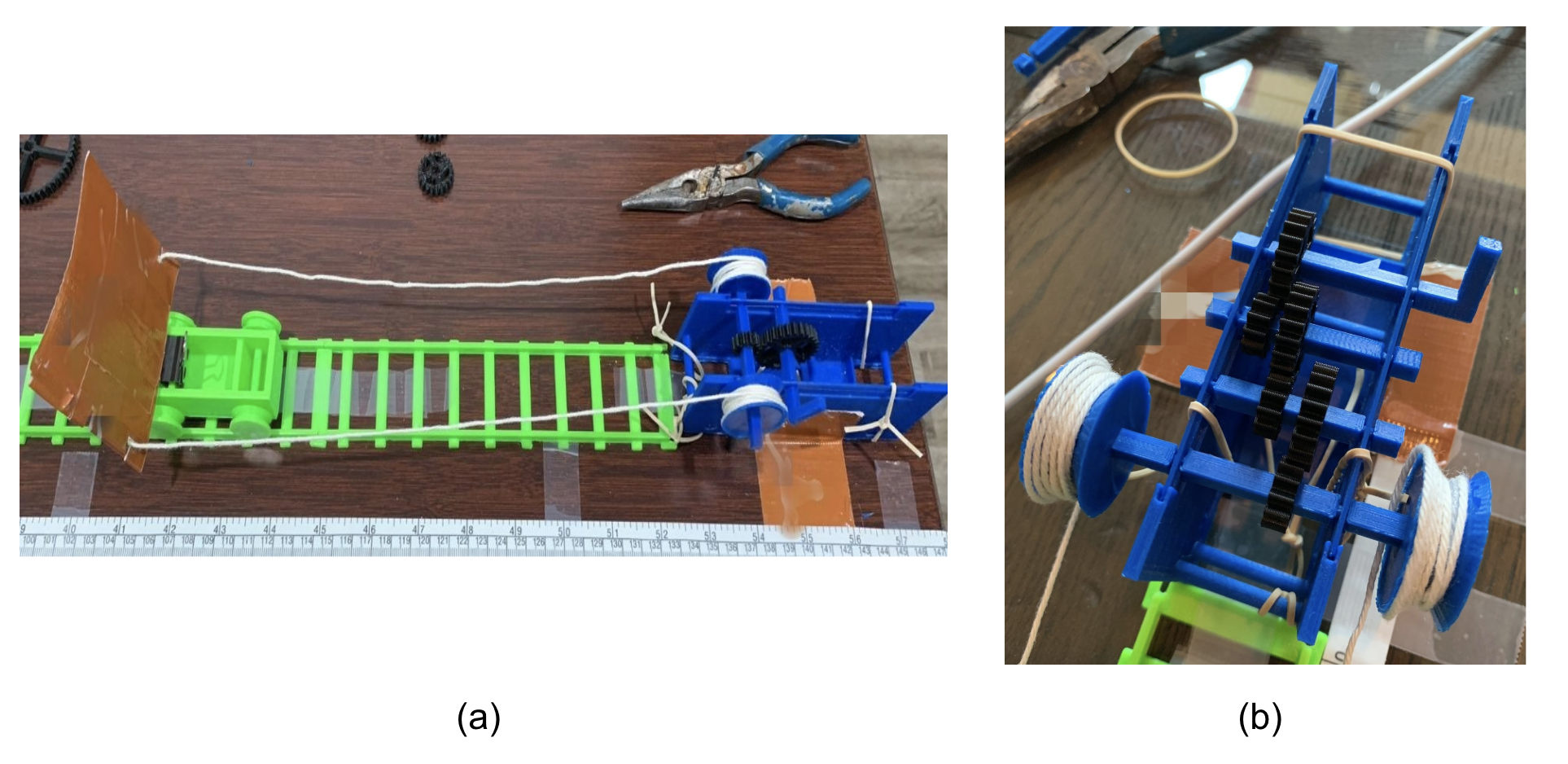}
\caption{(a) A top-view of the setup for physical experiments. The small car is pulled towards the gear box by the rope attached to the output axle and the traveled distance is recorded. (b) A closer visualization of the pulling mechanism.}
\label{fig_overall_evaluation}
\end{figure}

\begin{figure}
\includegraphics[width=2.0in]{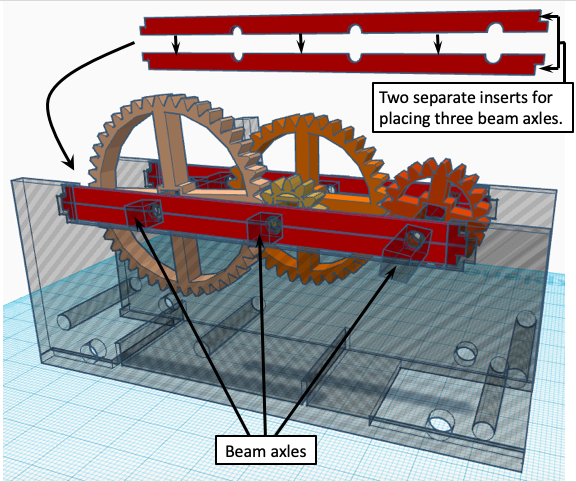}
\caption{Visualization of the pulling mechanism. The inserts (red parts, used to hold gears in place) are modified and re-printed for every design that is physically tested. All other components (gears, axles, and main body) are reused for each design.}
\label{cad}
\end{figure}

The overall test setup can be seen in Fig. \ref{fig_overall_evaluation}. The rail tracks, car being pulled and gear box are 3D printed. A rope, which wraps around the car, is connected to the output axle of the gear mechanism, located inside of the gear box. Torque is applied to the input axle of the gear mechanism by a rubber band that is twisted once around the axle. The axle is then released and spun by the rubber band, which causes the rest of the mechanism to be set in motion. The rotation of the output axle then wraps the rope around the pulleys on both ends of the output axle and pulls the car. The rail track is 35 inches long. The distance that the car travels along the track is recorded for each archived mechanism by the ruler located next to the rail track. The sail that is attached to the car is used to keep the rope out of the track so that it does affect the car's movement. On the other end of the track, a gear box is taped to the surface. This gear box, seen in Fig. \ref{fig_overall_evaluation}, was designed to be modular, such that the box and gears do not have to be 3D printed with each evaluation. Instead, only the inserts that hold the gear axles in place need to be printed (see Fig. \ref{cad}). The location of the holes within the inserts to hold axles in place are computed for each archive solution to allow the gears to mesh perfectly. This modular design greatly reduced the time required for testing each novel candidate mechanism.

Designing mechanisms that perform well in this environment is a difficult task. While a certain mechanism may have high output speed, its torque may not be enough to pull the car along the track because the weight of the car is not negligible. Because of the inverse correlation between the output speed and torque, each mechanism must find an effective balance between these two objectives to pull the car longer distances. In fact, many mechanisms having a high output speed and low torque, or vice versa, are only able to pull the car a few inches along the track. Because of such challenges, designing a gear mechanism to properly accomplish this task is not trivial. 


\section{Results}

\begin{figure}
\includegraphics[width=3.25in]{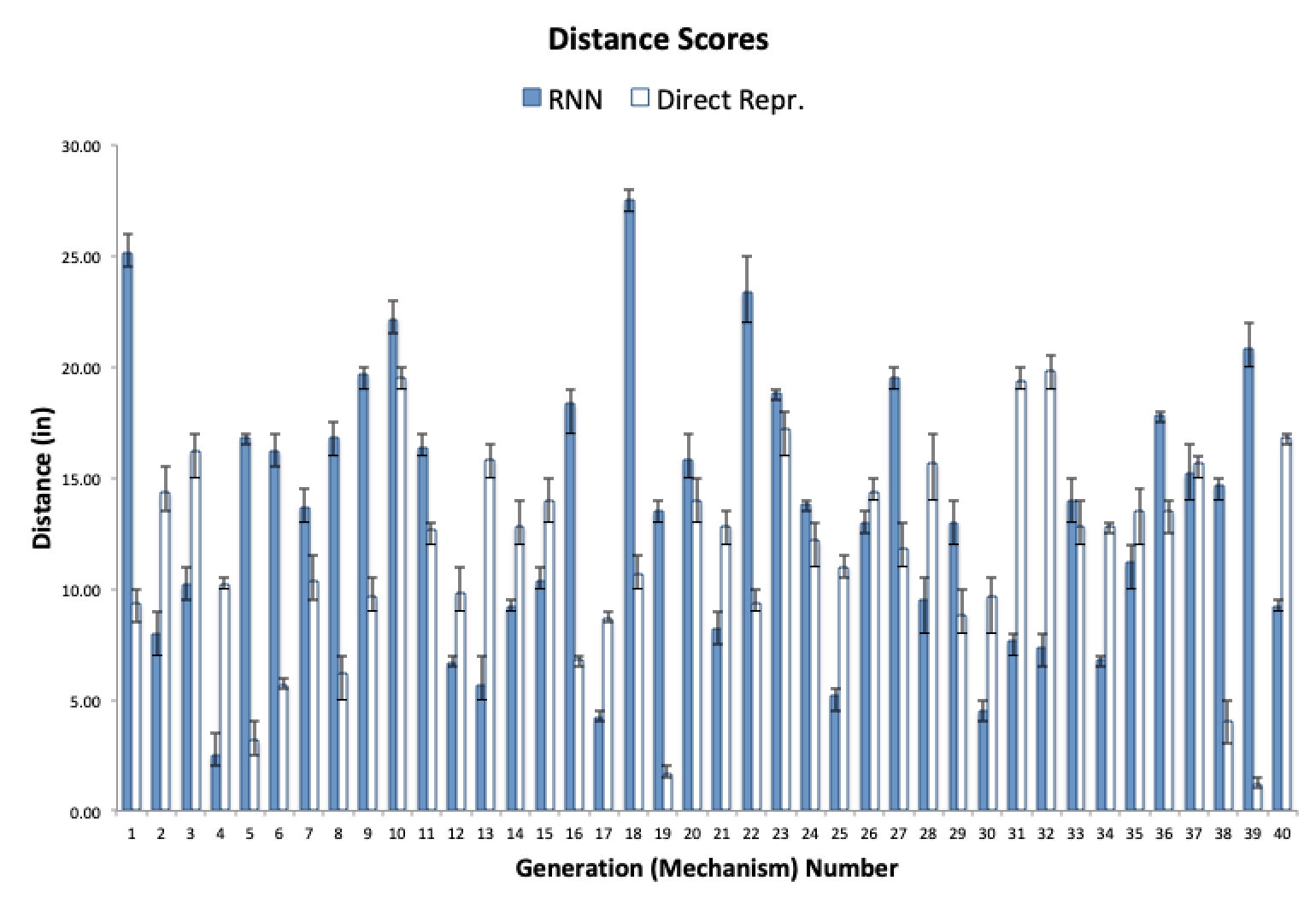}
\caption{Combined distance score results for all archived mechanisms generated by RNN and Direct Representation experiments. The error bars represent the maximum and minimum distances for the three tests that were performed for each mechanism, while the actual value reflects the average of the three trials.}
\label{mech_results}
\end{figure}

In this section, results for the evolution of RNN and direct representation models are presented. Two different experiments were performed to test the ability of the proposed generative RNN model and direct encodings to produce a diverse and functionally effective set of resulting mechanisms. The direct representation experiment is used as a baseline to determine the effectiveness of the generative RNN model. At each generation of evolution, the mechanism with the highest novelty was fabricated and tested to determine its distance score. Each mechanism was tested three times to ensure consistency and the distance score results were collected separately for RNN and direct representation experiments. Results for both methods are shown in Fig. \ref{mech_results} and discussed in detail in the next two subsections.

\subsection{Results of RNN Evolution}
\begin{figure}
\includegraphics[width=3.0in, height=1.25in]{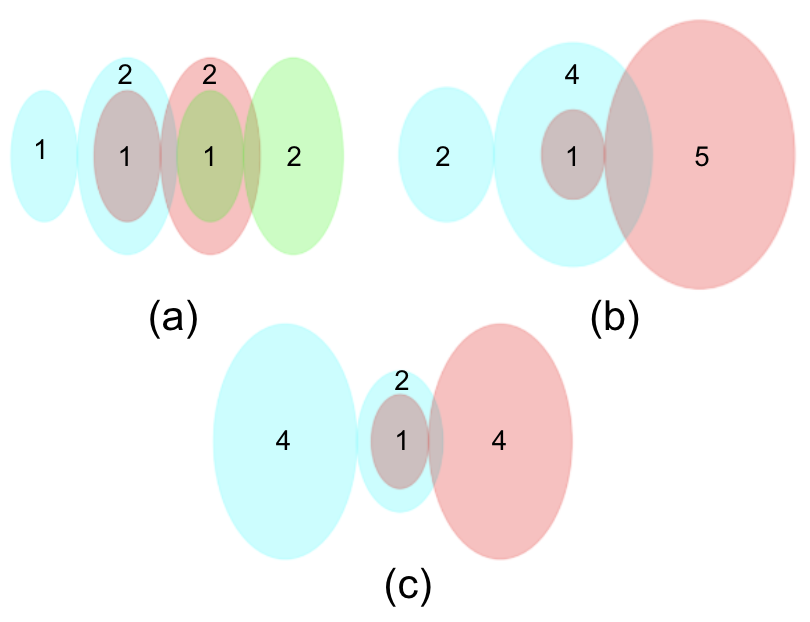}
\caption{Mechanism 1 (a), 18 (b), and 22 (c) produced by evolution of RNNs. Gears of the same color exist within the same plane and are connected linearly, while gears of different colors are coaxial. The numbers on each gear represent the gear type.}
\label{RNN_mechex}
\end{figure}

As seen in Fig. \ref{mech_results}, the generative RNN model managed to generate a diverse set of mechanisms that pulled the car varying distances. Among all mechanisms generated by RNN, the standard deviation of distance scores was 6.14 inches. In the final set, the three highest-performing mechanisms, each of which utilized a complex coaxial structure, pulled the car 27.5, 25.2, and 23.3 inches (mechanisms 18, 1, and 22, respectively). Their design patterns can be seen in Fig. \ref{RNN_mechex}. Out of 40 archived mechanisms, eight of them contained coaxial structure. Typically, such coaxial designs are difficult to create because they require a larger number of gears, which increases the probability of violating constraints. However, the RNN methodology was able to efficiently place gears into the system by utilizing repeating, coaxial design patterns that avoided collisions and minimized the size of the mechanism.

\begin{figure*}
  \includegraphics[width=\textwidth]{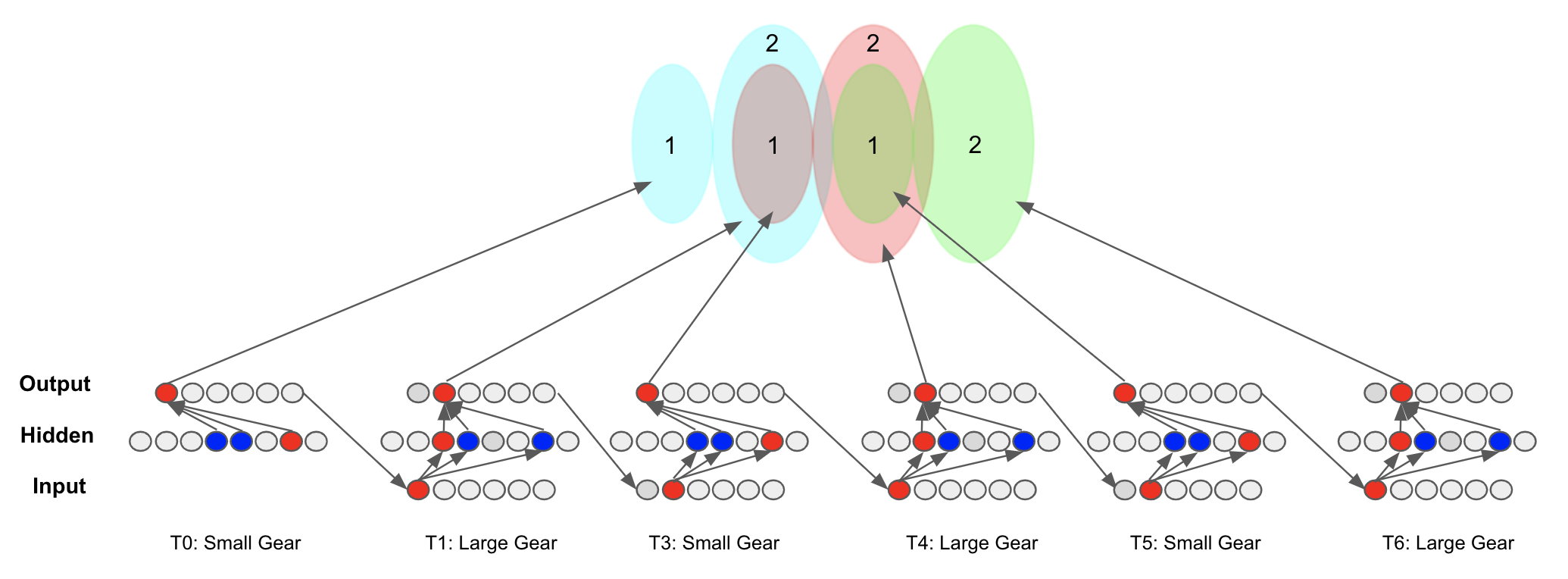}
  \caption{Visualization for design patterns found in the network structure for mechanism 1 from RNN evolution. The RNN memorizes the use of previous gears in the system to create sequential patterns in the output mechanism.}
\label{RNN_pattern}
\end{figure*}

When examining the design patterns in Fig. \ref{RNN_mechex}, it is clear that the generated mechanisms follow sequential, repetitive design patterns. For example, in mechanism (a), the pattern of placing a small gear adjacent to a large gear is repeated several times in the mechanism's structure. Such patterns are quite common in the resulting mechanisms from the evolution of RNNs. In Fig. \ref{RNN_pattern}, the composition of the RNN network that encodes mechanism (a) from Fig. \ref{RNN_mechex} is displayed. In this figure, the RNN's hidden state follows a distinct pattern while outputting gears at each time step. When a smaller gear is placed into the system, the hidden layer values contain a pattern that causes the next outputted gear to be large. Similarly, when a larger gear is placed into the system, the hidden layer values contain a pattern that causes the next outputted gear the be small. This pattern is repeated in every time step of the RNN's activation, thus revealing that the RNN utilizes the history of previously chosen gears to design the mechanism. In this case, it is clear that the RNN, through evolution, has learned a design rule for creating complex mechanisms, which causes it to always pair large gears with small gears. By following this design rule, the RNN creates an output mechanism that has a repeating, coaxial structure, which allows the mechanism, despite being complex, to be packed into a smaller area.
Such patterns were common to many of the RNNs that created complex structures, thus revealing RNN's ability to learn successful design patterns and use the history of previous gears to produce complex mechanisms without violating constraints. Such an ability to learn useful design techniques allowed RNN to focus evolution on a more efficient region of the search space. Because RNN is able produce unique mechanisms within the constraints of the design space, such mechanisms achieve relatively high novelty scores, allowing them to be explored, evolved, and physically tested.

\subsection{Results of Direct Representation Evolution}

\begin{figure}
\includegraphics[width=3.25in, height=1.05in]{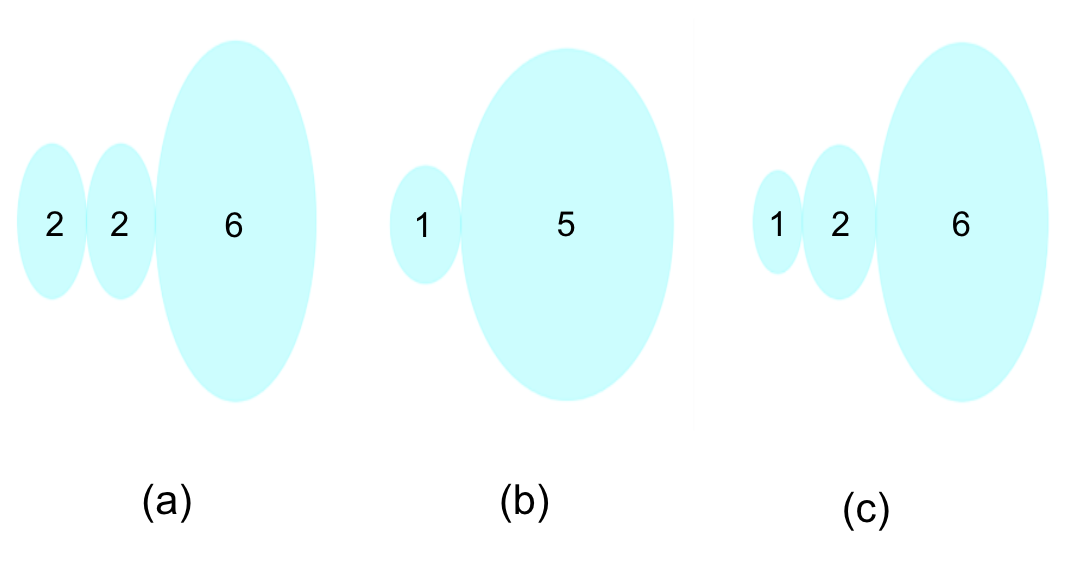}
\caption{Mechanisms 32 (a), 31 (b), and 10 (c) generated by evolution of direct representations. Gears of the same color are connected linearly.}
\label{GA_mechex}
\end{figure}

The evolution of direct representations also produced a set of feasible mechanisms. The overall standard deviation of distance scores is 4.6 inches. The three highest-performing mechanisms pulled the car 19.8, 19.5 and 19.3 inches (mechanisms 32, 12, and 31, respectively). These mechanisms are shown in Fig. \ref{GA_mechex}. Out of all 40 archived mechanisms, only three of them utilized coaxial gears.  None of these coaxial mechanisms performed nearly as well as those discovered by generative RNN models, which indicates that the evolution of direct representations was unsuccessful in finding effective coaxial design patterns. The rest of the mechanisms contained no coaxial gears in their structure.

\begin{figure}
\includegraphics[width=3.25in, height=1.05in]{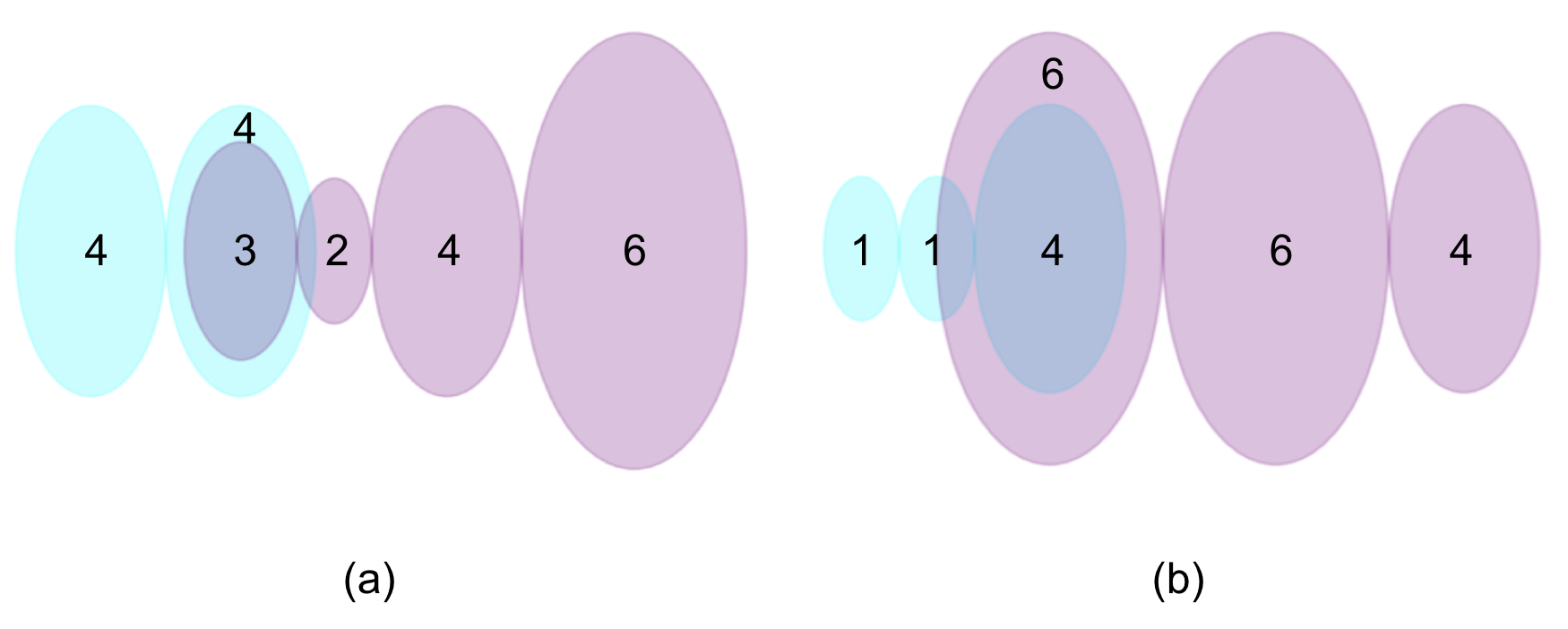}
\caption{Example of infeasible coaxial mechanisms generated by evolution of direct representations. The randomness and inefficiency of the designs can be observed. The numbers on each gear represent the gear type.}
\label{GA_badmech}
\end{figure}

The infeasible mechanisms generated in this experiment can be examined to better understand the difficulties with evolving complex and effective mechanisms, such as those discovered in the generative RNN experiment. As can be seen in Fig. \ref{GA_badmech}, coaxial design patterns found by the evolution of direct representations are comprised of random sets of gears, having no distinct design patterns. If these gears are not packed efficiently into the system, they cannot fit into the available design space and will violate constraints. Therefore, complex mechanisms evolve less efficiently due to the lack of feasible designs. Because the direct representation does not utilize history of previously selected gears in the mechanism, the model must find efficiently-packed design patterns through random search. Therefore, it is less efficient in learning the feasible design space for coaxial mechanisms.


\section{Discussion and Future Work}
The proposed generative RNN model is shown to be effective in evolving complex and feasible gear mechanisms despite the limited number of evaluations. Application of various genetic operators (selection, crossover and mutation) enabled RNNs to find a balance between complexity and lack of constraint violation, thus creating mechanisms that are more diverse and effective than those produced by the direct encoding. Despite the success of the proposed methodology, there are a couple of promising improvements to consider. First of all, automating the generative procedure with a realistic physics simulator would enable more detailed sampling and search in the design space. However, the use of sliced model (the object model used for fabrication in 3D printing) in such physics simulators is very limited due to the need for computational resources. The complexity of capturing the correct behavior of interacting parts (frictional surfaces, nonlinear material properties) is another limiting factor. Additionally, the generation of gears with continuous design parameters (i.e., the radius of the gear, the size and shape of gear teeth, etc.) with RNN could be explored.

\section{Conclusions}

This paper proposes a novel RNN model for generating a variety of functional 3D-printed gear mechanisms. The methodology consists of two main components: First, an RNN to represent gear mechanisms using indirect encoding. Second, novelty search to evolve more diverse and feasible designs that can be 3D printed and reliably work. When compared to a direct encoding of such gear mechanisms, the RNN model managed to produce more structurally diverse set of functional designs given a limited number of design evaluations. Lastly, the RNN was able to learn sequential design rules and was proved to consistently utilize such design rules to create more efficient mechanism structures.

\section{Acknowledgments}

The authors wish to acknowledge the funding and support provided
by the BEACON Research Center at Michigan State University.


\bibliographystyle{ACM-Reference-Format}
\bibliography{cite_vrs7} 

\end{document}